\documentclass[letterpaper, 10 pt, conference]{ieeeconf}  

\IEEEoverridecommandlockouts                              

\usepackage{cite}
\usepackage{amsmath,amssymb,amsfonts}
\usepackage{algorithmic}
\usepackage{graphicx}
\usepackage{textcomp}
\usepackage{cuted}
\usepackage{capt-of}
\usepackage{float}
\usepackage{booktabs}
\usepackage{threeparttable}
\usepackage{mathtools}
\usepackage[mathscr]{eucal}
\usepackage[table,xcdraw]{xcolor}
\usepackage{colortbl}
\usepackage{colortbl}
\usepackage{balance}
\usepackage{color}
\usepackage{multirow}
\usepackage{multicol}
\usepackage{marvosym}
\usepackage{algorithmic}
\usepackage{indentfirst}
\usepackage{makecell}
\usepackage{caption}

\makeatletter
\let\NAT@parse\undefined
\makeatother
\usepackage{hyperref}

\title{\LARGE \bf
Data Scaling Laws for Imitation Learning-Based End-to-End Autonomous Driving
}

\author{Yupeng Zheng$^{1,2,3}$,
Pengxuan Yang$^{1,2,3}$,
Zhongpu Xia$^{1}$\thanks{Project leader},
Qichao Zhang$^{1,2}$\thanks{Corresponding author},
Yuhang Zheng$^{4}$, 
Songen Gu$^{2}$, \\
Bu Jin$^{2}$,
Teng Zhang$^{3}$,
Ben Lu$^{3}$,
Chao Han$^{3}$,
Xianpeng Lang$^{3}$,
and Dongbin Zhao$^{1,2}$ \\
\textsuperscript{1}CASIA,
\textsuperscript{2}UCAS,
\textsuperscript{3}Li Auto,
\textsuperscript{4}NUS,\\
}

\begin{document}

\twocolumn[{%
\renewcommand\twocolumn[1][]{#1}%
\maketitle
\vspace{-8mm}
\begin{center}
    \captionsetup{type=figure}
    \includegraphics[width=0.93\textwidth]{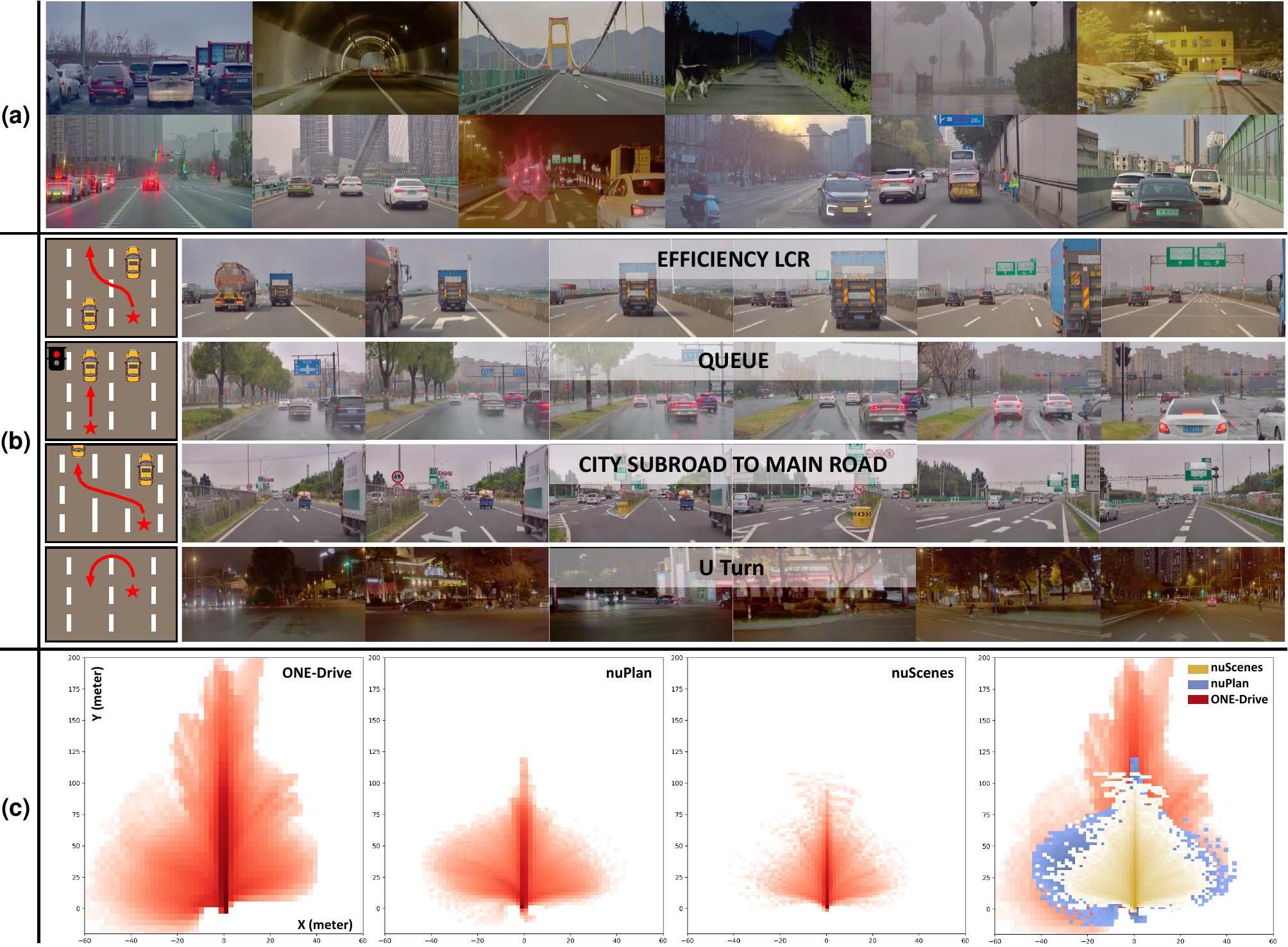}
    \centering
    \caption{\small{We have collected and utilized a large-scale, diverse real-world dataset, enabling us to investigate data scaling laws of end-to-end driving. Figure (a) illustrates the diversity of our dataset, encompassing various weather conditions, road types, and traffic scenarios. Figure (b) presents 23 scenario types that we have identified to conduct in-depth analyses of the impact of data scale on generalization and the importance of data distribution. Figure (c) compares the trajectory distributions of existing datasets nuScenes\cite{caesar2020nuscenes} and nuPlan\cite{caesar2021nuplan} with ours. Our trajectory distribution exhibits greater diversity, including a higher proportion of high-speed driving, turning, and lane changes.}}
    \label{fig:teaser}
\end{center}
\vspace{1mm}
}]


\begin{abstract}
The end-to-end autonomous driving paradigm has recently attracted lots of attention due to its scalability.
However, existing methods are constrained by the limited scale of real-world data, which hinders a comprehensive exploration of the scaling laws associated with end-to-end autonomous driving.
To address this issue, we collected substantial data from various driving scenarios and behaviors and conducted an extensive study on the scaling laws of existing imitation learning-based end-to-end autonomous driving paradigms.
Specifically, approximately 4 million demonstrations from 23 different scenario types were gathered, amounting to over 30,000 hours of driving demonstrations. We performed open-loop evaluations and closed-loop simulation evaluations in 1,400 diverse driving demonstrations (1,300 for open-loop and 100 for closed-loop) under stringent assessment conditions.
Through experimental analysis, we discovered that (1) the performance of the driving model exhibits a power-law relationship with the amount of data, but this is not the case in closed-loop evaluation. The inconsistency between the two assessments shifts our focus toward the distribution of data rather than merely expanding its volume.  (2) a small increase in the quantity of long-tailed data can significantly improve the performance for the corresponding scenarios; (3) appropriate scaling of data enables the model to achieve combinatorial generalization in novel scenes and actions.
Our results highlight the critical role of data scaling in improving the generalizability of models across diverse autonomous driving scenarios, assuring safe deployment in the real world.
\end{abstract}

\section{Introduction}
\label{sec:intro}
End-to-end autonomous driving has gained significant attention in recent years. 
It typically employs a differentiable model that takes raw sensor data as input and generates a potential planning trajectory as output. 
This paradigm allows for the direct optimization of the entire system in a data-driven manner, offering scalability where performance improvements can be achieved by increasing training data, as exemplified by scaling laws\cite{kaplan2020scaling, tian2024visual, henighan2020scaling}. 

Some previous works \cite{li2024hydra, anonymous2024drivetransformer} have tried to investigate the effect of scaling up.
However, they share the same challenge: insufficient real-world data, resulting in data scaling laws in end-to-end autonomous driving remaining under-explored.
As shown in the table.~\ref{tab:dataset}, the open-source datasets \cite{caesar2020nuscenes, caesar2021nuplan} are typically thousand-scale, far less than the million-scale or billion-scale vision-language data in language models or generative models. Although simulators like CARLA \cite{dosovitskiy2017carla} offer a promising solution to the data lack, the huge domain gap hinders their real-world application. 
Consequently, it remains under-explored whether end-to-end autonomous driving has such common data scaling laws and how autonomous driving vehicles can benefit from the laws. 

In this paper, we delve into the data scaling laws of end-to-end autonomous driving in the real world. 
We aim to investigate the three critical questions: 
\begin{itemize}
\item[$\bullet$] \textit{Is there a data scaling law in the field of end-to-end autonomous driving?}
\item[$\bullet$] \textit{How does data quantity influence model performance when scaling training data?}
\item[$\bullet$] \textit{Can data scaling endow autonomous driving cars with the generalization to new scenarios?}
\end{itemize}

To answer the questions, we collect and annotate a million-scale dataset named ONE-Drive,  which contains over 4 million driving demonstrations (about 30,000 hours) of real-world data in diverse cities and road conditions, as shown in the figure.~\ref{fig:teaser}. 
Based on the dataset, we first conduct a detailed analysis of the relationship between training data and model performance. 
Further, we split the scenarios into 23 types based on the traffic condition and agent behavior to delve into the generalization of the data scaling law. 
Lastly, we set the training data with different distributions on scenario types to analyze the relationship between data scaling and data distribution. 
For a comprehensive analysis, we conduct the open-loop test, closed-loop test, and real-world deployment to evaluate the planning results.

Throughout our experiments, we surprisingly find: 
\begin{itemize}
\item[$\bullet$] \textbf{Scaling law.} The data scaling law in open-loop evaluation exhibits a power-law relationship with the amount of data, but this is not the case in closed-loop evaluation. The inconsistency between the two evaluations shifts our focus toward the distribution of data rather than merely expanding its volume.  (Section \ref{sec:scaling}).
\item[$\bullet$] \textbf{Importance of increasing the quantity of targeted data.} Targeted augmentation of long-tailed scenarios exhibits superior efficiency compared to indiscriminate expansion of the overall dataset.  (Section \ref{sec:distribution}).
\item[$\bullet$] \textbf{Combinatorial generalization.} With the scaling of training data, models gradually acquire combinatorial generalization capabilities, enabling them to combine known information to achieve generalizable planning for new scenarios and actions (Section \ref{sec:generlization}).

\end{itemize}

\begin{table}[ht]
\caption{Comparison between the previous dataset and our introduced ONE-Drive for the end-to-end autonomous driving task.}
\vspace{-1mm}
\centering
\resizebox{1\linewidth}{!}{
\begin{tabular}{l|ccccc}
\toprule
\multirow{2}{*}{Dataset} & \multicolumn{2}{c}{Setting} & \multirow{2}{*}{Source} & Scale \\
\cmidrule(lr){2-3}
& Open-loop & Visual closed-loop & & (demonstrations)  \\ 
\midrule
Bench2drive~\cite{jia2024bench2drive} &  & {\checkmark} & Simulator & 10k \\
nuScenes~\cite{caesar2020nuscenes} & {\checkmark} &  & Real-world & 1k  \\
nuPlan~\cite{caesar2021nuplan} & {\checkmark} &  & Real-world & ~20k \\
ONE-Drive (Ours) & {\checkmark} & {\checkmark} & Real-world & \textbf{~4M} \\
\bottomrule
\end{tabular}
}
\label{tab:dataset}
\vspace{-2mm}
\end{table}

\section{Related Work}
\label{sec:related_work}

\begin{figure*}
  \centering
  \includegraphics[width=0.95\textwidth]{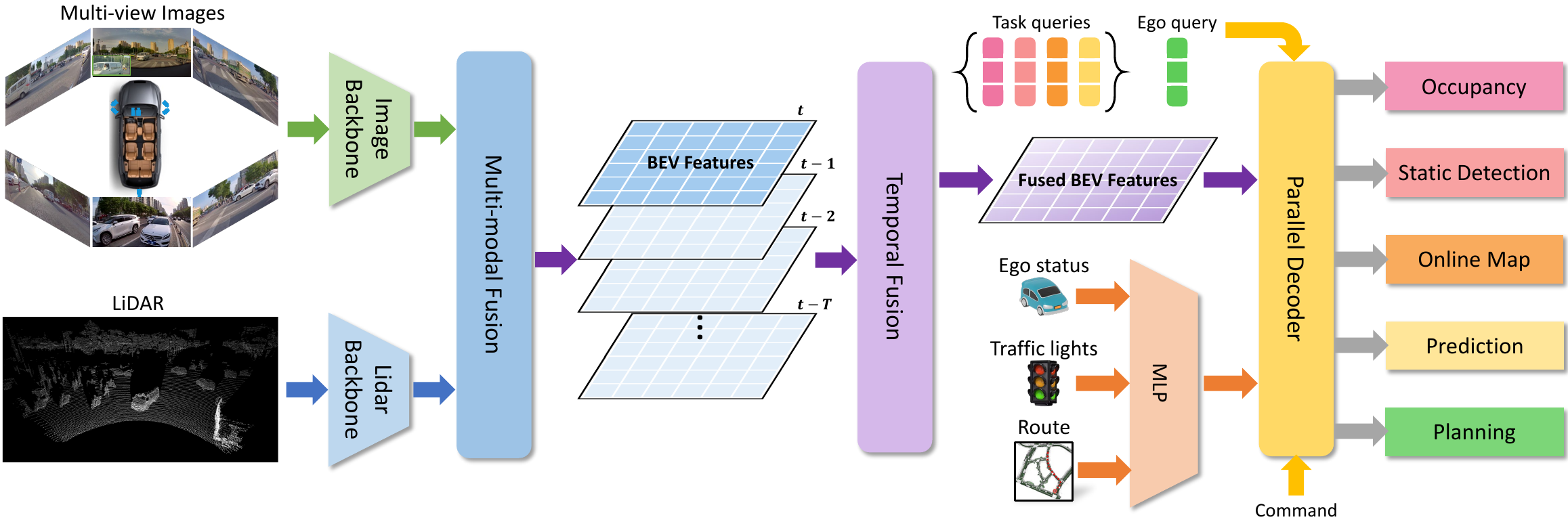}
  \caption{\small To better investigate the data scaling laws of end-to-end algorithms, we implement a parallel baseline methodology. This approach, inspired by the PARA-Drive~\cite{weng2024drive} framework, offers enhanced training stability and computational efficiency.}
  \label{fig:main}
\end{figure*}

\subsection{End-to-end Autonomous Driving}
 
Recent years have witnessed the development of autonomous driving vehicles \cite{hu2023planning, jiang2023vad, anonymous2024drivetransformer, chitta2022transfuser, li2024hydra, sun2024sparsedrive, shao2023safety}. Among them, end-to-end approaches catch great attention for their direct optimization of the whole system, offering a potential solution for the information loss and cascade error of traditional module-based approaches. They typically take raw sensor data as input and generate a potential path or plan as output. 
Some previous works focus on end-to-end solutions in simulators like CARLA \cite{dosovitskiy2017carla}. For example, TransFuser \cite{chitta2022transfuser} and InterFuser \cite{shao2023safety} combine information from different sensors to enhance the robustness and performance of the autonomous driving model in complex scenarios by fusing perception and planning. TCP \cite{wu2022trajectory} demonstrates exceptional performance using only a monocular camera by introducing target trajectory-guided control prediction, surpassing other methods that use multi-sensor inputs (multi-camera and LiDAR). AD-MLP \cite{zhai2023rethinking} presents a simple method based on a multilayer perceptron (MLP) that directly outputs the future trajectory of the vehicle from raw sensor data input.
However, the domain gap between the simulator and the real world hinders their industrial practice. To avoid the domain gap, some work \cite{hu2023planning} tries end-to-end autonomous planning in the real world. For example, UniAD \cite{hu2023planning} utilizes BEV queries to unify different tasks with the transformer, and VAD \cite{jiang2023vad} proposes to represent the scene in a vectorized space. More recently, BEVPlanner \cite{li2024ego} proposes a novel metric that enables a more comprehensive evaluation of model performance. 
Such data-driven optimization enables the ability to improve the system by simply scaling training resources. 
The end-to-end approaches offer a potential solution for the information loss and cascade error of traditional module-based approaches. 

\subsection{Scaling Laws}

In recent years, the development of foundation models has highlighted the significance of scaling laws in the field of large language models\cite{kaplan2020scaling, henighan2020scaling}, vision generative models\cite{peebles2023scalable, zhai2022scaling, henighan2020scaling, alayrac2022flamingo, dong2023dreamllm, liu2024neural}, and robotics\cite{lin2024data, zhao2024aloha, khazatsky2024droid, o2023open, bharadhwaj2024roboagent, walke2023bridgedata, naumann2025data}. These laws elucidate the relationship between dataset size, model size, and performance, showing that the effectiveness of transformer-based models follows a power-law function relative to the number of model parameters, the volume of training data, and the computational resources used for training. This concept has also been extended to other domains, such as image and video synthesis\cite{tian2024visual}, allowing for a more efficient trade-off between resource allocation and performance outcomes. However, there is limited research on the application of scaling laws to end-to-end autonomous driving, primarily due to constraints related to data availability. It still remains uncertain whether any scaling law exists that links the scalability of neural networks to planning results in autonomous driving, which is the main focus of our work.

\section{Method}\label{sec:method}

Following PARA-Drive~\cite{weng2024drive}, we implement a parallel modular end-to-end autonomous driving algorithm named the ONE model, as shown in the figure.~\ref{fig:main}. The input is LiDAR point clouds and panoramic RGB images, and the expected output is a planning trajectory. 

\subsection{BEV Encoder}
Given the input of multi-view images and point clouds, we generate a unified Bird's Eye View (BEV) representation through the View Transformation module and the Temporal Fusion module. 

\textbf{View Transformation.} 
For multi-view camera images, we utilize an image backbone to extract image features. Then we utilize LSS \cite{philion2020lift} to project these image features into a 3D frustum form. For LiDAR point clouds, we extract voxelized LiDAR features with 3D sparse convolution layers. The features are then flattened along the height dimension, leading to features in a BEV plane. Finally, the image features and point cloud features are fused with squeeze-and-excitation (SE) blocks \cite{hu2018squeeze}.

\textbf{Temporal Fusion.} 
To effectively leverage historical information, we employ a temporal fusion module. We set up a FIFO queue to sequentially extract and store past BEV features and relative poses. During training and inference, the features stored in the queue are transformed using the pose information to align with the current pose. The transformed features are then fused with the current BEV features by SE blocks \cite{hu2018squeeze}. Finally, the queue is updated by adding the current frame features to the end of the queue and dequeuing the front element of the queue.

\begin{figure*}[!t]
  \centering
  \includegraphics[width=0.85\textwidth]{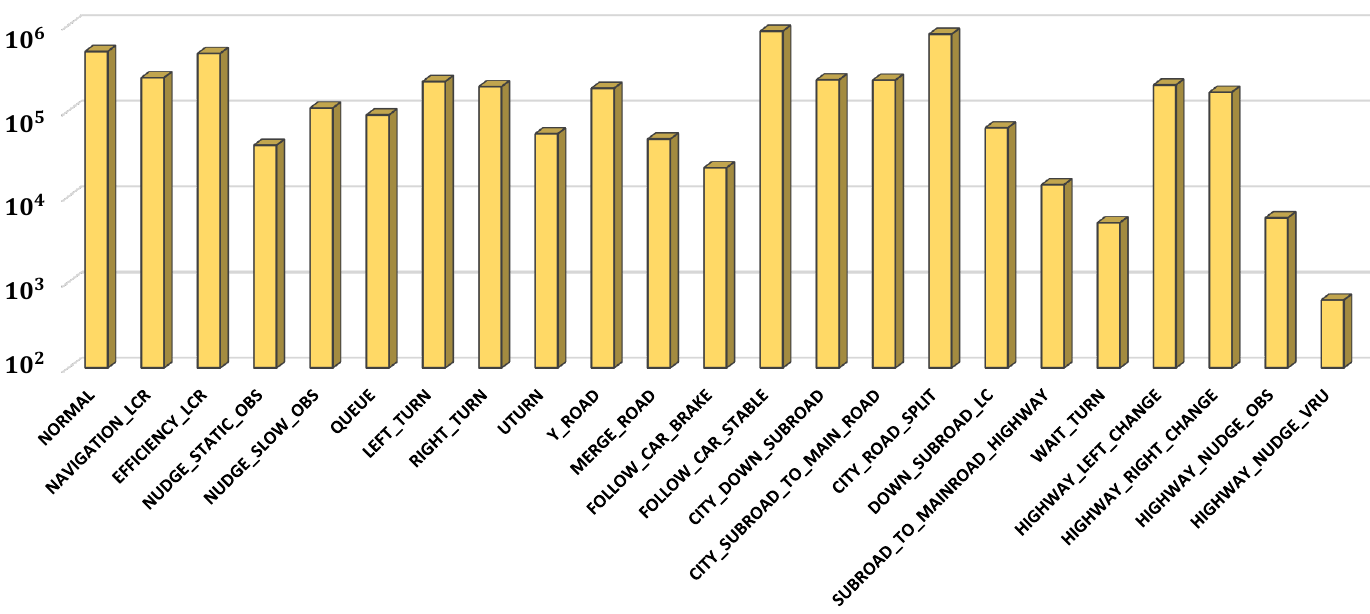}
  \vspace{-2mm}
  \caption{\small Distribution of the 23 scenario types.}
  \label{fig:scenario}
\end{figure*}

\subsection{Parallel Decoder}
Our decoder comprises \textbf{five tasks}: online map prediction, 4D occupancy prediction, static detection, motion prediction, and planning. Inspired by \cite{hu2023planning, jiang2023vad, sun2024sparsedrive}, we design each module with learnable query features tailored specifically for its corresponding task.

\textbf{Online Mapping. }
For online map prediction, we utilize a set of map queries to estimate a vector map from BEV features, along with the class scores corresponding to each map vector. The road map labels include lane dividers, lane centerlines, crosswalks, and stop lines. We use the loss function defined in MapTR \cite{liao2022maptr}.

\textbf{4D Occupancy Prediction. } 
For occupancy, we establish a set of occupancy queries to estimate the 3D occupancy and the corresponding flow for the next 3 seconds from BEV features. The 3D occupancy prediction loss is the same as ~\cite{cao2022monoscene} and the flow loss employed in Cam4DOcc~\cite{ma2024cam4docc}.

\textbf{Static Detection.} 
We specifically design an additional static object detection task to address the challenge of capturing small static objects in occupancy prediction. We employ Hungarian matching to match the query features with ground truth values.

\textbf{Motion Prediction.} 
For dynamic objects, we follow \cite{jiang2023vad} by setting a set of agent queries. Each agent query interacts with the BEV features through attention layers to capture environmental information. Similarly, we supervise the motion prediction of dynamic objects using the L1 loss.

\textbf{Ego Motion Planning.} 
In the planning module, to fulfill the requirements of real-world planning for precise adherence to navigation, we employ a multi-layer perceptron (MLP) to encode the traffic light status, road-level route (represented by a collection of dense waypoints at the road level), and the motion state of the ego-vehicle, which serve as contextual information for planning. Following \cite{jiang2023vad, weng2024drive}, we leverage learnable embeddings to construct ego-vehicle queries. These queries are passed as input to the cross-attention layers to interact with BEV features. Then the queries are used to predict multi-modal future planning trajectories with corresponding scores.  
During inference, we execute the trajectory with the highest confidence score.

\begin{figure*}
  \centering
  \includegraphics[width=0.96\textwidth]{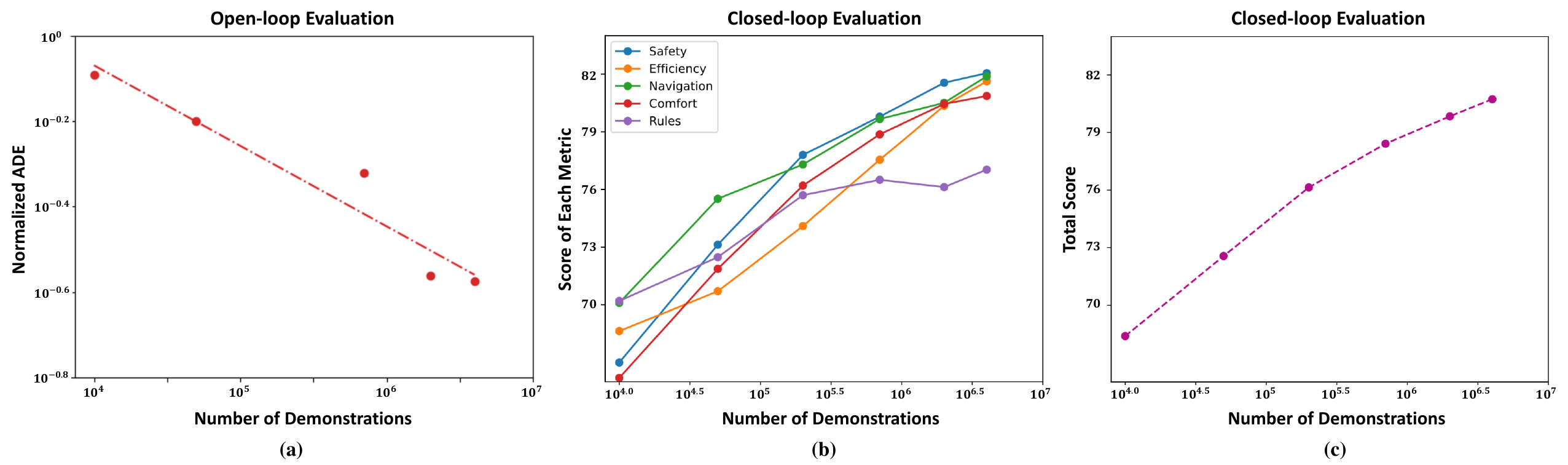}
  \caption{\small Results of open-loop [(a)] and closed-loop [(b) and (c)] evaluation. \textbf{(a):} Data scaling laws on ADE.  Dashed lines represent power-law fits, with the equations and coefficient $r$ provided in the eq.~\ref{eq:power}. All axes are shown on a logarithmic scale. \textbf{(b):} Scores of each metric in closed-loop evaluation. \textbf{(c):} Total score of closed-loop evaluation. In (b) and (c), the X-axis (number of demonstrations) is displayed on a logarithmic scale, and the Y-axis is displayed on a linear scale.}
  \label{fig:result}
\end{figure*}

\section{Experiments Setup}
\label{sec:exp_setup}
\subsection{Implementation Details}
\textbf{Data Collection.} 
The scene images were captured horizontally by seven cameras covering a 360° field of view (FOV), with each image having a resolution of $3840 \times 2160$. Additionally, we employed a 128-beam LiDAR sensor to acquire point cloud of the environment.

\textbf{Baseline.} 
Our method predicts a 6-second future trajectory with 2 seconds of historical information. 
It uses ResNet34~\cite{he2016deep} as the default image backbone. 
The image resolution is downsampled to $512 \times 960$ for training. 
The default BEV resolution is $232 \times 80$ for a perception range of $139.2m \times 48m$ longitudinally and laterally. 
The default hidden state and embedding dimension are 256.
We construct five training datasets with varying numbers of demonstrations: \textbf{10 thousand}, \textbf{50 thousand}, \textbf{0.2 million}, \textbf{0.7 million}, \textbf{2 million}, and \textbf{4 million}. We train each model to full convergence during training for a fair comparison. This allows models trained on different dataset sizes to undergo different numbers of training steps. Training steps and computational resources are shown in the table. \ref{tab:step}
\begin{table}[!t]
\centering
\renewcommand\arraystretch{1.1}
\caption{Training steps and resources of different training dataset sizes (demonstrations).}
\vspace{-1.5mm}
\resizebox{1\linewidth}{!}{
\setlength{\tabcolsep}{4pt}
\begin{tabular}{c|cc}
\toprule
Dataset Size (Demonstrations)   & Training Resource   & Training Steps  \\
\midrule
10 thousand & 64 A100 & $6.0 \times 10^5 $ \\ 
50 thousand & 64 A100 & $8.6 \times 10^5 $ \\
0.2 million & 128 A100 & $2.9 \times 10^6 $ \\
0.7 million & 320 A100 & $9.3 \times 10^6 $ \\ 
2 million & 512 A100 & $2.2 \times 10^7 $    \\ 
4 million & 512 A100 & $3.1 \times 10^7 $   \\ 
\bottomrule
\end{tabular}
\label{tab:step}
}
\end{table}

\subsection{Data Mining}\label{sec:mining}
To facilitate more flexible adjustment of data distribution, we define 23 scenario types according to the agent behavior and traffic conditions. They can be acquired based on the following meta-information:
\begin{itemize}
\item \textbf{Navigation Information}: This includes upcoming road names, road types, number of lanes, and distance to intersections.
\item \textbf{Static Perception}: This includes lane markings, driveable areas, intersection areas, and pedestrian walkways.
\item \textbf{Dynamic Perception}: This covers road obstacle information, including position, velocity, dimensions, historical trajectories, and predicted future trajectories.
\item \textbf{Ego Vehicle Status}: This comprises the ego vehicle's position, velocity, acceleration, historical trajectory, and future trajectory.
\end{itemize}
The distribution of these scenario types in the training dataset (4 million demonstrations) is shown in the figure. \ref{fig:scenario}.
For more details about the specific characteristics of the 23 scenario types, please see the supplementary materials.

\subsection{Evaluation}\label{sec:eval}
To comprehensively analyze the scaling laws of data, we evaluate our model using three methodologies: open-loop evaluation and closed-loop simulation.
It is important to acknowledge that, owing to constraints in the reconstruction process, the closed-loop evaluation environment has not been fully aligned with the open-loop.
Future work will align the two evaluation settings to make a fair comparison and analysis of the data scaling law in these two settings.

\textbf{Open-loop Evaluation.} The open-loop evaluation computes the Average Displacement Error (ADE) metric between the predicted trajectories and the ground truth trajectories, which can evaluate the network's ability to fit the ground truth trajectory.

\textbf{Closed-loop Simulation.} For closed-loop simulation, we employ 3D-GS~\cite{kerbl20233d} to reconstruct partial test scenarios, enabling end-to-end simulation with visual closed-loop feedback. This simulation evaluates the driving trajectory in five dimensions: \textbf{safety}, \textbf{comfort}, \textbf{rule}, \textbf{efficiency}, and \textbf{navigation}. 
For each metric, a higher score indicates better performance.
Finally, the total score is weighted by these five scores.
\begin{equation*}
\begin{split}
    score = & 0.25 \times \rm{safety} + 0.15 \times \rm{comfort} + 0.2 \times \rm{rule} \\
    &+ 0.25 \times \rm{efficiency} + 0.15 \times \rm{navigation}
\end{split}
\end{equation*}

\section{Experiments}
In this section, we investigate the relationship between the scale of training data and the ability of end-to-end driving models to fit expert trajectories (Section \ref{sec:power}). Subsequently, we discover the inconsistency in data scaling laws between open-loop and closed-loop evaluations, revealing that merely increasing data volume does not indefinitely improve autonomous driving planning performance in real-world scenarios (Section \ref{sec:diff}). This finding motivates us to explore data scaling strategies beyond simply augmenting data quantity. We examine how data distribution affects the efficiency of the data scaling law (Section \ref{sec:distribution}). Finally, we demonstrate that end-to-end autonomous driving systems, when scaled appropriately with data, exhibit combinatorial generalization capabilities.
\subsection{Unveiling of Data Scaling Laws}\label{sec:scaling}

\subsubsection{Power-law Relationship}\label{sec:power}
Inspired by the power-law data scaling laws observed in large-scale language models~\cite{kaplan2020scaling}, we aim to investigate the existence of power-law scaling relationships in imitation learning-based end-to-end autonomous driving. 
We begin our analysis by examining the Average Displacement Error (ADE) between ground truth trajectories and predicted trajectories in different training dataset scales.
Specifically, we utilize open-loop evaluation results from five models with varying training dataset scales, as detailed in Section \ref{sec:exp_setup}. 
We designate the amount of training data (quantified by the number of demonstrations used in training) as a variable $\rm{X}$ and the normalized trajectory ADE as a variable $\rm{Y}$. 
After applying the logarithmic transformation to both $\rm{X}$ and $\rm{X}$, the presence of a linear relationship would indicate a power-law scaling law in end-to-end autonomous driving data.

As shown in the figure.~\ref{fig:result} (a), we conduct the linear model fitting of the log-transformed data. The fitting results yield:
\begin{equation}\label{eq:power}
    Y = 0.6833 \cdot X^{-0.188}, r = -0.963
\end{equation}
The correlation coefficient $r$ of the fit is -0.963, strongly suggesting the existence of a power-law data scaling law in trajectory fitting for imitation learning-based end-to-end autonomous driving.

\subsubsection{Differences Between Open-loop And Closed-loop}\label{sec:diff}
Given that open-loop evaluation cannot fully reflect a vehicle's planning performance in real-world scenarios, we further assess the data scaling laws in a closed-loop condition. 
Specifically, we employ a visual closed-loop simulator to evaluate the vehicle's closed-loop performance of the models trained on variant scales of data. 
Following the description in Sec \ref{sec:eval}, we obtain the five closed-loop scores and the total score as illustrated in the figure.~\ref{fig:result} (b) and (c).

From figure (b), we observe distinct relationships between various closed-loop evaluation metrics and the expansion of data volume.
Metrics such as navigation and efficiency demonstrate a clear and stable increasing trend, as they are directly related to trajectory and path training. 
In contrast, the score for the driving rule, while showing an overall upward trend, exhibits instability during its growth. 
This occurs because driving rules are not explicitly modeled as loss functions in imitation learning—instead, the model must implicitly learn the patterns from large volumes of image-trajectory pairs.
Due to inherent limitations in imitation learning, such as causal confusion, performance in rule compliance remains suboptimal.
Similarly, metrics including safety and comfort experience rapid improvement initially, but their growth rates gradually slow within the range of 200k to 700k data samples. 
We hypothesize that in the early stage, increasing data volume enhances the model’s ability to fit trajectories, leading to swift gains in closed-loop safety. 
In later stages, however, minor discrepancies in trajectory fitting have a diminishing impact on safety in most scenarios. 
Thus, further improving safety may depend more on optimizing data distribution to enhance performance across diverse situations, rather than simply increasing the amount of data.

\subsection{Data Quantity and Model Performance}\label{sec:distribution}
Based on the foundation of the data scaling law, we explore the relationship between data quantity and model performance when increasing the scale of training data. Specifically, we categorized our dataset of 2 million demonstrations into 23 scenario types, as defined in Section \ref{sec:mining}, and analyzed the proportion of each type. 
To simplify the process, we identify two scenario types that exhibit poor performance (DOWN\_SUBROAD\_LC and SUBROAD\_TO\_MAINROAD\_HIGHWAY) from the 23 categories for in-depth study. By incrementally increasing the representation of these scenario types, we observe the consequent effects on planning performance.

As illustrated in the table. \ref{tab:distribution}, we maintained a constant overall volume of training data while increasing the quantity of these two scenario types in each iteration. Subsequently, we evaluated the model's open-loop trajectory error after training.
The results reveal that for these long-tailed scenarios, doubling the specific training data while keeping the total data volume constant led to improvements in planning performance ranging from 9.7\% to 16.9\%. Furthermore, quadrupling the specific training data yielded even more substantial gains, with performance improvements between 22.8\% and 32.9\%. 
Our findings suggest that even a relatively small increase in scenario-specific data (comprising only several hundred or thousands of demonstrations) can lead to significant enhancements in planning performance for these challenging scenarios.

\subsection{Combinatorial Generalization}\label{sec:generlization}
\begin{figure}
  \centering
  \includegraphics[width=0.46\textwidth]{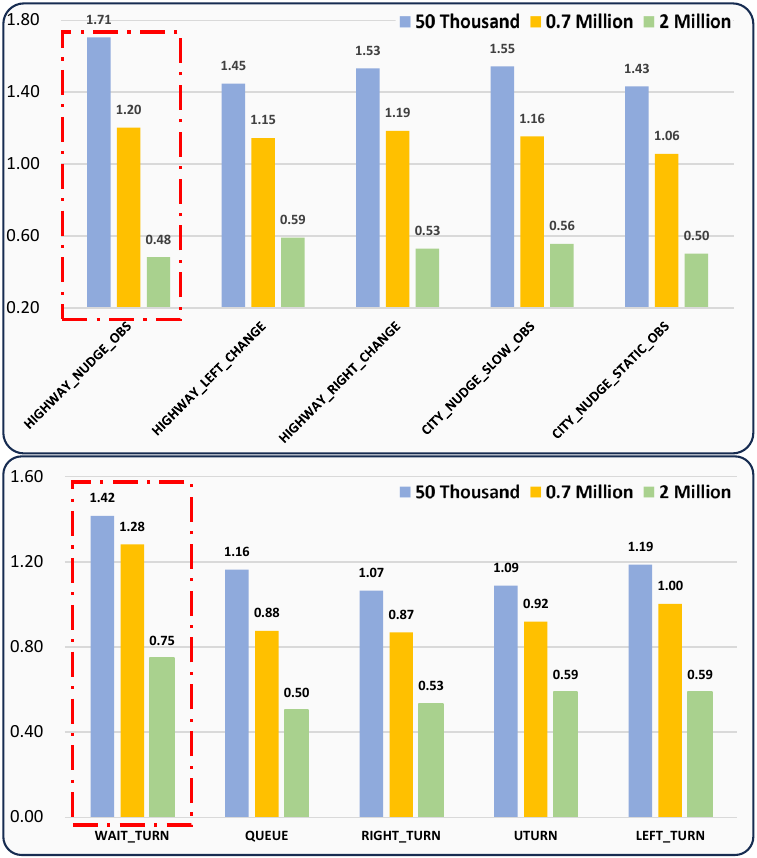}
  \caption{\small To investigate the relationship between data scaling and generalization ability, we select data from 2 scenario types, HIGHWAY\_NUDGE\_OBS and WAIT\_TURN as test data. Models trained with \textcolor[RGB]{143,170,220}{\textbf{50 thousand}}, \textcolor[RGB]{255,192,0}{\textbf{0.7 million}}, and \textcolor[RGB]{169,209,142}{\textbf{2 million}} demonstrations are represented in \textcolor[RGB]{143,170,220}{\textbf{blue}}, \textcolor[RGB]{255,192,0}{\textbf{yellow}}, and \textcolor[RGB]{169,209,142}{\textbf{green}}, respectively.} 
  \label{fig:generlization1}
\end{figure}
In this section, we investigate the relationship between data scaling laws and the generalization capability, which is considered essential for deploying autonomous driving in real-world scenarios.

\begin{table*}[ht]
\centering
\caption{Analysis of different scene distribution.}
\vspace{-1.6mm}
\resizebox{2.04\columnwidth}{!}{
\begin{tabular}{c|ccc|ccc}
\toprule
Scenario Type & \multicolumn{3}{c}{Data Quantity} & \multicolumn{3}{|c}{ADE $\downarrow$}  \\
\midrule
DOWN\_SUBROAD\_LC  & 4972 & 7218 ($+45.1\%$) & 21836 ($+339.2\%$) & 0.788  & 0.711 ($-9.7\%$)  & 0.529 ($-32.9\%$) \\ 
SUBROAD\_TO\_MAINROAD\_HIGHWAY  & 643 & 1485 ($+130.9\%$) & 2569 ($+299.5\%$) & 0.786  & 0.653 ($-16.9\%$)  & 0.593 ($-22.8\%$) \\ 
\bottomrule
\end{tabular}
}
\vspace{1mm}
\label{tab:distribution}
\end{table*}
\subsubsection{Quantitative Analysis}
To demonstrate the existence of generalization, we split two categories (HIGHWAY\_NUDGE\_OBS and WAIT\_TURN) from the 23 scenario types defined in Sec \ref{sec:mining} as test categories, while using the remaining 21 categories to form the training set. 
To be consistent with previous settings, we trained models using three scales of number of training data: \textbf{50 thousand}, \textbf{0.7 million}, and \textbf{2 million} demonstrations. 
Notably, we applied a strict filtering strategy when selecting test data for these two categories to ensure that each scenario does not overlap with other scenario types. 
The open-loop test results are shown in the figure.~\ref{fig:generlization1}. For a convenient comparison, we select scenarios similar to these two scenario types. 

We observed that (1) the model trained on \textcolor[RGB]{143,170,220}{\textbf{50 thousand}} demonstrations showed larger displacement error from expert trajectories in the two test scenarios compared to similar scenarios, indicating insufficient generalization capability with small-scale training data. (2) As the training data gradually increased to \textcolor[RGB]{169,209,142}{\textbf{green}} demonstrations, the displacement error between the trajectories of the two test scenarios and other scenarios rapidly narrowed. The performance on HIGHWAY\_NUDGE\_OBS even surpasses the performance on other scenarios that participated in the training. (3) By learning high-speed driving and low-speed nudging obstacles separately from the training data, the model acquired the ability to generalize to high-speed HIGHWAY\_NUDGE\_OBS scenarios; through learning to turn and queuing at red lights, the model developed the capability to generalize to WAIT\_TURN scenarios.
Based on these observations, we argue that \textbf{the data scaling law endows models with the ability to perform combinatorial generalization.}

\begin{figure*}
  \centering
  \includegraphics[width=0.9\textwidth]{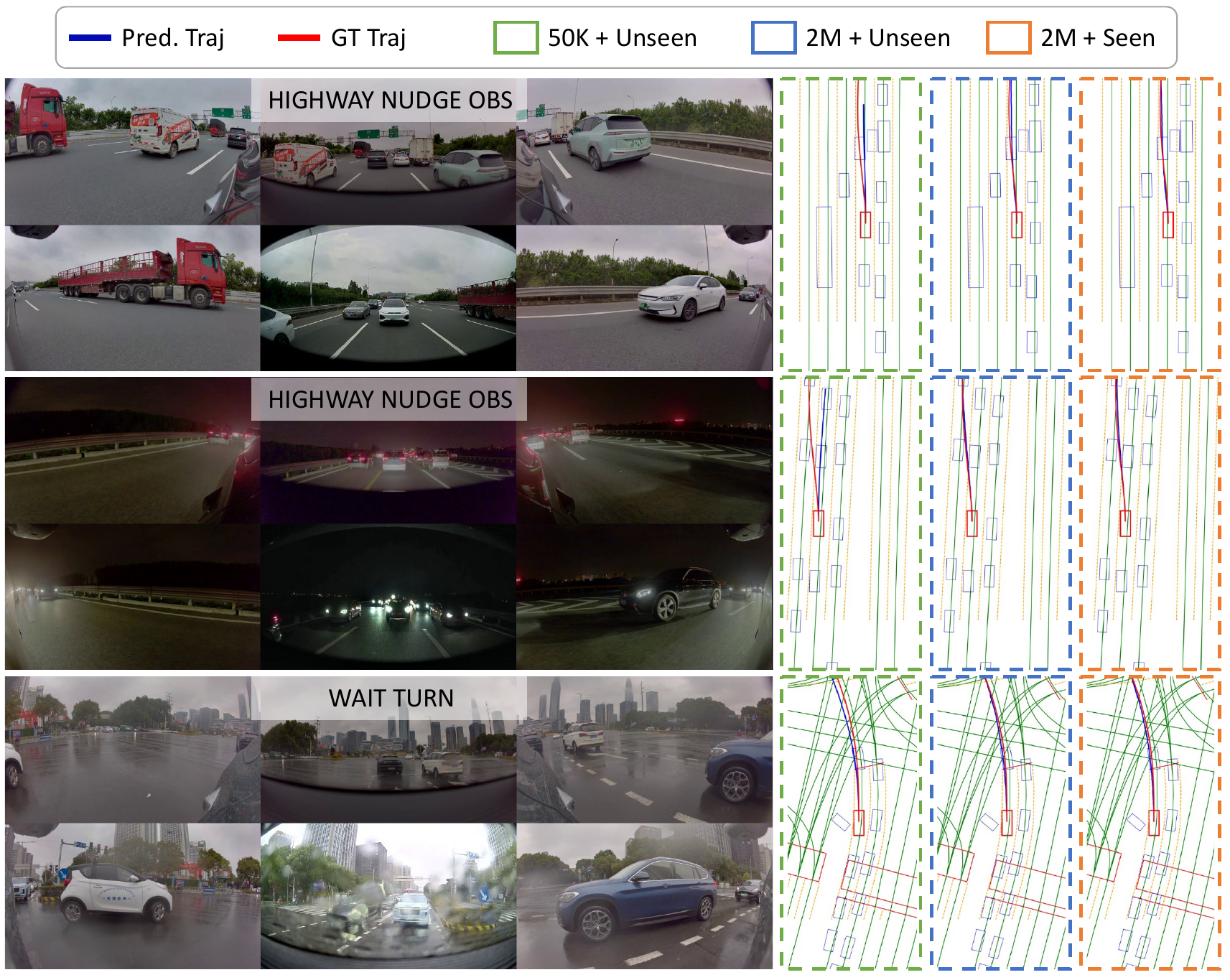}
  \vspace{-2mm}
  \caption{\small Qualitative results on scenarios of HIGHWAY\_NUDGE\_OBS and WAIT\_TURN scenario types. We compare the planning trajectories generated by three different models, demonstrating the combinatorial generalization in new driving scenarios.}
  \vspace{-0.5mm} 
  \label{fig:vis}
\end{figure*}

\subsubsection{Qualitative Analysis}
To demonstrate the generalization ability more intuitively, we also provide a qualitative visualization of the predicted trajectory and ground truth map in the figure. \ref{fig:vis}, which shows the planning results of HIGHWAY\_NUDGE\_OBS and WAIT\_TURN scenarios.
The BEV maps in the \textcolor[RGB]{129,172,84}{\textbf{green boxes}} represent the planning result of the model trained using 50 thousand demonstrations without HIGHWAY\_NUDGE\_OBS and WAIT\_TURN scenarios (\textit{50K + Unseen} in the legend), the BEV maps in the \textcolor[RGB]{87,113,191}{\textbf{blue boxes}} represent the planning result of the model trained using 2 million demonstrations without HIGHWAY\_NUDGE\_OBS and WAIT\_TURN scenarios (\textit{2M + Unseen} in the legend), and the BEV maps in the \textcolor{orange}{\textbf{orange boxes}} represent the planning result of the model trained using 2 million demonstrations with HIGHWAY\_NUDGE\_OBS and WAIT\_TURN scenarios (\textit{2M + Seen} in the legend).

Visual analysis reveals that appropriately increasing the scale of training data enables the model to achieve combinatorial generalization to novel scenarios. This enhanced generalization capability allows the model to perform competitively with counterparts specifically trained in these new scenarios. Our findings underscore the critical role of data scaling in improving model adaptability and robustness across diverse autonomous driving contexts.

\section{Conclusion}
\subsection{Conclusion}
In this paper, we delve into the data scaling laws of the imitation learning-based end-to-end autonomous driving framework. Upon further investigation, we uncovered three intriguing findings: \begin{itemize}
    \item A power-law data scaling law in open-loop metric but different in closed-loop metric.\vspace{1mm}
    \item Data distribution plays a key role in the data scaling law of end-to-end autonomous driving.\vspace{1mm}
    \item The data scaling law endows the model with combinatorial generalization, which powers the model's zero-shot ability for new scenarios. \vspace{1mm}
\end{itemize}

\section{Limitation and Future Work}
While our work provides significant insights, we acknowledge several limitations. 
Due to computational constraints, our investigation primarily focused on imitation learning methods based on mean regression with Bird's Eye View (BEV) representation. 
To address these limitations and further enrich our understanding of data scaling laws in autonomous driving, we propose the following directions for future research:
\begin{itemize}
    \item exploring scaling laws for a broader range of model architecture (visuomotor policies and vision-language-action);
    \item investigating how self-supervised pre-training changes the scaling law; 
\end{itemize}
These avenues could provide valuable insights into the scalability and generalizability of autonomous driving systems across diverse scenarios and model paradigms.

\balance
\bibliographystyle{IEEEtran}
\bibliography{ref}
\end{document}